\documentclass[a4paper, 10pt, conference]{ieeeconf}      

\IEEEoverridecommandlockouts                              

\title{\LARGE \bf
Traffic-Domain Video Question Answering with Automatic Captioning
}

\author{
Ehsan Qasemi$^{1}$ and Jonathan M. Francis$^{2}$ and Alessandro Oltramari$^{2}$
\thanks{$^{1}$Department of Computer Science, University of Southern California, Los Angeles, CA, USA
        {\tt\small qasemi@usc.edu}}%
\thanks{$^{2}$Human-Machine Collaboration, Bosch Center for Artificial Intelligence, Pittsburgh, PA, USA
        {\tt\small \{jon.francis, alessandro.oltramari\}@us.bosch.com}}%
}

\usepackage{graphicx}
\usepackage{amsmath}
\usepackage{booktabs}
\usepackage{multirow}
\usepackage{hyperref}
\usepackage{comment}
\usepackage{xspace}
\usepackage{float}

\newcommand{\QT}[1]{``{#1}''\xspace}

\newcommand{\ITMo}{\texttt{ITMo}\xspace}

\newcommand{\papername}{\texttt{TRIVIA}\xspace}
\newcommand{\tqa}{\texttt{SUTD-TrafficQA}\xspace}
\newcommand{\citet}[1]{\citename{#1}}

\begin{document}

\maketitle
\thispagestyle{empty}
\pagestyle{empty}

\begin{abstract}

Video Question Answering (VidQA) exhibits remarkable potential in facilitating advanced machine reasoning capabilities within the domains of Intelligent Traffic Monitoring and Intelligent Transportation Systems. 
Nevertheless, the integration of urban traffic scene knowledge into VidQA systems has received limited attention in previous research endeavors. 
In this work, we present a novel approach termed Traffic-domain Video Question Answering with Automatic Captioning (\papername), which serves as a weak-supervision technique for infusing traffic-domain knowledge into large video-language models. 
Empirical findings obtained from the \tqa~task highlight the substantial enhancements achieved by \papername, elevating the accuracy of representative video-language models by a remarkable 6.5 points (19.88\%) compared to baseline settings. This pioneering methodology holds great promise for driving advancements in the field, inspiring researchers and practitioners alike to unlock the full potential of emerging video-language models in traffic-related applications.

\end{abstract}

\section{Introduction}
\label{sec:introduction}
Intelligent Traffic Monitoring (\ITMo) represents an essential instrument to improve road safety and security in intelligent transportation systems~\cite{pascale2012wireless,9112118,Hutson2018-hl}, estimated to be worth approximately \$3B\@ in the U.S.~\cite{GrandView21}.
With further adoption of autonomous vehicles, \ITMo is poised to become an even more relevant part of the smart city infrastructure of the future~\cite{chowdhury2021towards,nees2016acceptance,duarte2019self}.

The main challenge in \ITMo is related to the fusion of large quantities of multi-modal information (e.g., surveillance cameras feed, acoustic sensors) and the integration of state-of-the-art (SOTA) reasoners to utilize such knowledge \cite{hans}.
Multiple studies have addressed different aspects of \ITMo, either by proposing symbolic methods for reasoning over and managing multimodal sensor data~\cite{10.1145/3314344.3332480,hans,muppalla2017knowledge,lam2017real,souleyrette2003remote}, proposing knowledge resources related to the domain~\cite{trafficqa,Chowdhury2021TowardsLC}.
Specifically, \tqa~\cite{trafficqa} is a crowdsourced video question-answering resource that includes 6 reasoning tasks corresponding to various traffic scenarios. 
SOTA's performance on these tasks remains far below human-level performance.

With the advent of transformer-based language and video-language models \cite{li2022uniformer,zellers2021merlot,fu2021violet}, which are SOTA on the \tqa~task, there has been a widespread push to find methods to inject domain-specific knowledge into such models~\cite{lin2019kagnet,ma2019towards-hykas,ma2021knowledge,sap2019atomic,qasemi2022pinks,trafficqa}.
\begin{figure*}
    \centering
    \includegraphics[width=\textwidth]{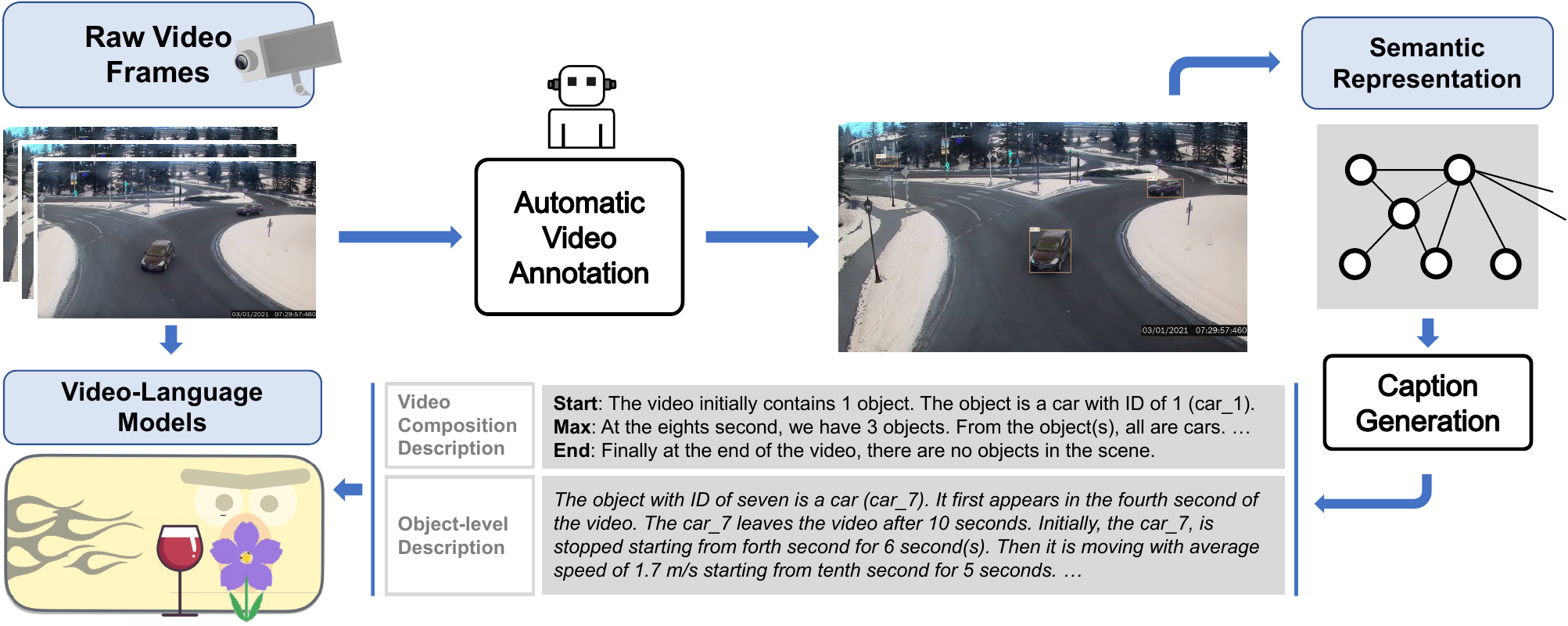}
    \caption{Overview of \papername for urban traffic knowledge-injection for Video Question Answering.}
    \label{fig:overivew}
\end{figure*}
In this work, we propose a method for \textit{\textbf{Tr}affic-doma\textbf{i}n \textbf{Vi}deo Question-Answering with \textbf{A}utomatic Captioning} (\papername)---a weak supervision method to inject traffic-domain knowledge into video-language models.
\papername starts from raw urban traffic videos, widely available online, for different geographic locations.\footnote{e.g.,
\hyperlink{NY511}{NY511}, YouTube, crowd-curated resources such as \tqa~\cite{trafficqa}, etc.}
We use off-the-shelf video/image annotations tools (for example \cite{zhou2022global,fischer2022qdtrack,yolo,hou2019vehicle,Singh2022VIPHYPC,carion2020detr}) to extract information such as objects, their position, color, trajectory, and speed from these traffic scenes, then
we use the HANS framework~\cite{hans} to store and manage the information in a symbolic knowledge graph.
To infuse the SOTA video-language model reasoners, we propose a 
\textit{synthetic captioning} method to inject traffic domain knowledge, inspired by the common training process of video-language models: we use automatic methods to generate synthetic natural language descriptions (captions) for each video based on the above extracted information.
The resulting video-caption pairs are used to fine-tune the model to inject it with the traffic domain information.
We show how our synthetic captioning method can improve the accuracy of the VIOLET video-language model \cite{fu2021violet} on \tqa~\cite{trafficqa} task, by 6.5 percentage points, through curriculum learning \cite{pentina2015curriculum}.

\section{Background and Related Works}
\label{sec:background}
In this section, we briefly cover the backgrounds of video language models and the utilization of weak supervision methods to improve language models across different natural language understating spectrums.

\subsection{Video Language Models}
A broad spectrum of downstream tasks, from question-answering\cite{rajpurkar2016squad} to event forecast\cite{ahuja2019language2pose} can benefit from video language models(\cite{li2022uniformer,zellers2021merlot,fu2021violet}), which can ground reasoning on the interconnection between vision and language. 
From the implementation standpoint, video language models borrow the notion of transformer architecture\cite{vaswani2017attention,liu2019roberta} from language models, and realize different variations of attention across modalities.
For example, VIOLET \cite{fu2021violet} has two pipelines for encoding visual and textual information into embedded features: it then uses a Cross-modal Transformer to combine the video and text features into unified contextualized embeddings.
For the video pipeline, VIOLET uses the video swine-transformer \cite{liu2022video} to model sparsely sampled video frames along both spatial and temporal dimensions as video features.
For the text pipeline, VIOLET follows WordPiece \cite{wu2016google}, where word tokens are considered as textual features.

\subsection{Weak Supervision for Knowledge Injection}

In weak supervision, the objective closely resembles that of supervised learning. 
However, instead of relying on human experts to directly annotate unlabeled data, an alternative approach involves leveraging expert knowledge to create user-defined patterns that can be used to infer \QT{noisy} or \QT{imperfect} labels ~\cite{rekatsinas2017holoclean,zhang2017deepdive,dehghani2017neural,Singh2022VIPHYPC}.
Obtaining such imperfect labels can be achieved through the application of heuristic rules or by re-purposing external knowledge~\cite{alfonseca2012pattern,bunescu2007learning,mintz2009distant}, as well as other forms of domain knowledge~\cite{stewart2017label}.
Weak supervision has found extensive utility in the field of Natural Language Understanding (NLU).
Notable applications include its use in enhancing the understanding of object affordances~\cite{qasemi2022pinks}, extracting temporal commonsense information from raw text~\cite{zhou2020temporal}, generating reasoning rationale~\cite{brahman2020learning}, improving neural ranking models~\cite{dehghani2017neural}, and enhancing translation in African languages\cite{hedderich2020transfer}.
Furthermore, weak supervision has been applied in the development of ASER~\cite{zhang2020aser} and ASCENT~\cite{nguyen2020advanced}, which are frameworks utilized to extract relations from unstructured text.

Language models can be considered implicit knowledge bases that store a vast amount of information about the world.
Consequently, querying language models as a source of weak supervision represents a feasible approach. 
For instance, authors in \cite{wang-etal-2021-table-based} employ language model-based augmentation to enhance the saliency of data in tables, authors in \cite{meng-etal-2021-distantly} employ language models as a source of weak supervision in named entity recognition, and authors in \cite{dai2021ultra} utilize masked language models for weak supervision in entity typing.

\section{Methodology}
\label{sec:methodology}
The schematic representation of the \papername system can be observed in Fig. \ref{fig:overivew}. 
This section provides a comprehensive explanation of our meticulous approaches employed in the acquisition of videos, generation of automatic annotations, production of automatic captions for each video, and finally, using the automatically generated parallel video-captions to inject traffic domain knowledge into a video language model.

\subsection{Restrictions on the Input Videos}

The raw traffic data utilized in \papername consists of video feeds obtained from stationary cameras, such as traffic cameras. 
We deliberately exclude ego-view video\footnote{for example, from the vehicle's perspective} sources like A2D2~\cite{geyer2020a2d2}, Argoverse~\cite{chang2019argoverse}, and CAP-DATA~\cite{fang2022cognitive} due to the inherent difficulties associated with obtaining reliable automatic annotations from such sources. 
Notably, the variation in perspectives within ego-view videos can negatively impact the accuracy of speed predictions, making them unsuitable for our purposes.

\subsection{Video Annotations}

\papername employs readily available off-the-shelf automatic video/image annotation tools to generate annotations, acknowledging the potential presence of noise in the generated annotations. 
Object detectors are utilized to identify objects of interest within each frame while tracking algorithms are employed to assign persistent IDs to these detected objects across video frames. 
It is important to note that the scope of the study is limited to urban traffic scenarios, resulting in the detection of six specific classes: car, truck, bus, motorbike, pedestrian, and bicycle. 
To estimate the trajectory and size of the objects, bounding box size is utilized as a noisy approximation of object size. 
Furthermore, for color extraction, the model proposed in VIPHY\cite{Singh2022VIPHYPC} is applied to automatically extract the top color candidates for each detected bounding box.
 
\subsection{Semantic Representation}

HANS\cite{hans} is a neuro-symbolic architecture and framework for multi-modal context understanding for \ITMo. 
It utilizes knowledge graph technology to serve as a backbone and proposes an ontology for traffic monitoring (\textit{domain level}).
HANS is built to extend the Scene Ontology (introduced in \cite{tiddi2020neuro}; \textit{core level}), which extended W3C's Semantic Sensor Network ontology (SSN) specifications (introduced in \cite{haller2019modular}; \textit{foundational level}). 
HANS employs appropriate mapping mechanisms to instantiate the domain ontology using video annotations, thereby generating a comprehensive Traffic Monitoring Knowledge Graph. 

We chose to utilize HANS instead of the newly-designed ASAM's OpenX standard\cite{openx} primarily due to HANS's intuitive nature and widespread adoption in modeling temporal properties.
Unlike OpenX, which associates temporal extensions directly with participants, the SSN ontology represents time instants and intervals as properties of events. 
For example in OpenX, the relations \QT{hasBeginning} and \QT{hasEnding} are predicated over instances of the type \textit{car}. 
The conceptual choice in SSN aligns better with the nature of camera-based feeds commonly encountered in traffic monitoring scenarios. 
For instance, a single instance of type \textit{car} can participate in multiple events within a video, each event having its distinct start and end times. 
Furthermore, OpenX lacks direct definitions for temporal quantities, reducing them to the \QT{PhysicalQuantity} property with floating-point values. This modeling approach hampers temporal reasoning, which necessitates the presence of \textit{time stamps} and derived time intervals as \textit{first class citizens} in the ontology of reference.

\subsection{Automatic Caption Generation}
To generate the captions for the extracted semantic knowledge, we employ a series of template sentences, as a widely used method for lexicalizing structured knowledge (for example \cite{ma2019towards-hykas,bouraoui2020inducing}). 
These templates, such as \QT{The video initially contains <count> <object>,} are populated with information derived from the annotations stored in the HANS ontology.

Automatic captions generated in \papername consist of two main sections/paragraphs: composition, and features.

The composition paragraph provides an overview of the video's general composition, including the number of objects and their respective counts (as depicted by the top greyed box in Fig.~\ref{fig:overivew}). 
To maintain conciseness, we focus on four specific moments within the video: the beginning, the frame with the highest number of objects, the frame with the lowest number of objects, and the end.

The features paragraph delves into the characteristics of individual objects within the video. 
Each sentence in this section explores features such as object type, speed, motion direction, appearance time in the scene, persistence length in the scene, disappearance time from the scene, and more. 
To avoid an excessive number of sentences arising from transient features like speed and direction, we select salient changes as checkpoints and report average values for the periods in between. 
This approach effectively captures the notion that changes in direction and accelerations hold valuable information for scene understanding tasks.

\subsection{Traffic-Domain Knowledge Injection}
One effective strategy for injecting domain-specific knowledge into language models, and by extension video language models, is through the process of fine-tuning them on data specific to that domain (for example as used in \cite{liu2019roberta}). 
By fine-tuning a pre-trained model on a corpus of parallel text and video that is relevant to the target domain, in our case traffic domain, the model can learn to better understand language specific to that domain. 
This process allows the model to capture the nuances, terminology, and patterns of the domain, ultimately enhancing its performance and effectiveness in tasks within that specific domain, thereby providing a valuable means to incorporate and leverage domain-specific knowledge.

Hence, as the final step in our framework, we fine-tune the Video-LM model on the combination of raw traffic video and automatically-generated textual captions.

\section{Evaluation}
\label{sec:evaluation}
In our framework, we employ a novel approach to enhance the performance of general video language models (VidLMs) by injecting video annotations-derived captions into the models. This infusion of traffic knowledge aims to improve their efficacy in downstream tasks. In this section, we elaborate on the experimental setup utilized and examine the impact of the proposed \papername setup on the overall performance of a video-LM in a representative traffic-domain video question-answering task.

\subsection{Dataset}

\begin{figure}
    \centering
    \includegraphics[width=.8\columnwidth]{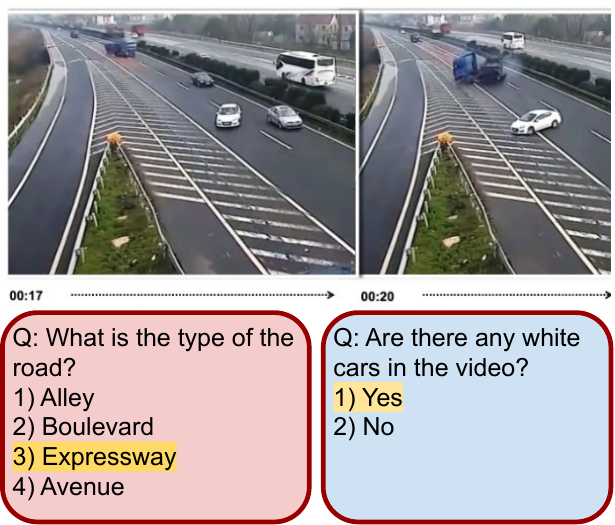}
    \caption{An example from the video-based question-answer pairs in \tqa~\cite{trafficqa}.}
    \label{fig:tqa_example}
    \vspace{-1.0em}
\end{figure}

We utilize \tqa~\cite{trafficqa} as our target task, which pertains to the domain of video question-answering in a multiple-choice format. 
An illustrative example is presented in Fig. \ref{fig:tqa_example}. 
The primary objective of \tqa is to evaluate models in comprehending traffic scenarios across various levels of complexity. 
This entails fundamental recognition aspects such as determining the type of road (through the question \textit{what is the type of the road?}), as well as more sophisticated reasoning abilities like counterfactual inference (e.g., assessing the likelihood of an accident occurring with fewer vehicles through the question \textit{would the accident still happen if there were fewer vehicles?}) and event forecasting (e.g., predicting whether a white sedan will collide with a barrier through the question \textit{will the white sedan crash into the barrier?}).

The \tqa dataset encompasses 10,080 real-world videos and relies on human annotators to generate a total of 62,535 question-answer pairs. 
Among these pairs, 56,460 are included in the training set, while the remaining 6,075 constitute the test set. 
It is noteworthy that 33,522 QA pairs pertain to stationary video perspectives, such as those obtained from traffic cameras, whereas 29,011 pairs involve an ego-view perspective, typically captured by autonomous vehicles. 
This dataset serves as the basis for evaluating the performance of the proposed \papername framework on a general video language model.
As the source of raw videos for \papername, utilized the videos made available by \tqa and derived automatic captions from these video sources. Our training and tuning process strictly excluded any information sourced from \tqa beyond the raw video material.

\subsection{Implementation Details}

We employ the HANS framework~\cite{hans} for the purpose of storing and managing information within a symbolic knowledge graph. HANS is implemented utilizing Stardog\cite{stardog}, a widely recognized knowledge graph framework designed for enterprise-level applications.
To perform object detection and tracking within the video data, we utilize a proprietary object tracker that has been specifically fine-tuned for traffic scenarios. 
This enables us to accurately extract objects and track their movements throughout the duration of the video.
To maintain an optimal balance between scene granularity, storage capacity, and computational query time, we process the videos at a rate of 1 frame per second. 
This approach aligns with our engineering requirements and ensures efficient handling of the data.

\begin{table}[ht]
\caption{Stats on average detected objects in videos reflected in the generated captions.}
    \centering
    \footnotesize
    \begin{tabular}{|l|c|}
        \hline
        Object Type & Average Count \\
        \hline
        Car & 18.23\\
        Pedestrian & 1.30 \\
        Motorbike & 0.08 \\
        Truck & 0.78 \\
        Bicycle & 0.08 \\
        Bus & 0.09 \\
        \hline
        Total & 20.57 \\
        \hline
    \end{tabular}
    \label{tab:det-stats}
\end{table}

\subsection{Experimental Setup}

We employed the VIOLET video language model \cite{fu2021violet} (refer to Section~\ref{sec:background}), which represents a state-of-the-art (SOTA) model with the ability to simultaneously process video and text.
To utilize the VIOLET model for the multiple-choice question-answering task in \tqa, we introduced a linear layer (dense layer) as the classification head on top of the embeddings generated by VIOLET.
The classification head is responsible for producing the label corresponding to the correct choice, and it is a commonly adopted practice for transformer-based architectures applied to classification tasks \cite{liu2019roberta}.
We compare the label generated for the choice with the correct value and report the model's accuracy on the test portion of the \tqa in two setups.

The results are presented and compared for two setups.
Firstly, we assess the baseline performance of VIOLET on the test subset after fine-tuning it on the training subset of \tqa, using cross-entropy loss.
In the second setup, VIOLET undergoes a traffic knowledge injection process and is subsequently fine-tuned on the training subset of \tqa, followed by evaluation on the test subset.
For the knowledge injection process, as discussed in Section \ref{sec:methodology}, we employ the generated captions along with the original video (for which the captions are generated) as training data, fine-tuning the VIOLET model based on its original loss function: Masked Visual-token Modeling (MVM) \cite{fu2021violet}.
To accommodate the model's size to fit on standard hardware, we freeze the parameters in the language encoder and the video encoder (swine transformer) within the VIOLET architecture.
Given that the VIOLET model operates on video frames extracted from the input video, according to its default configuration, we sample 5 frames from the video for use in the remaining stages of the pipeline.

\subsection{Results}
\paragraph{Statistics of Generated Captions:}

The statistical analysis reveals that the average length of the generated captions is 5556.14 characters, encompassing approximately 996.39 words and comprised of an average of 67.25 sentences. These captions exhibit a vocabulary of 219 distinct words, excluding numerical representations. The pertinent information pertaining to the extracted objects, which is directly manifested in the caption generation, is concisely presented in Table \ref{tab:det-stats}.

\paragraph{Performance on \tqa:}
The performance of the baseline model, i.e. the VIOLET model without traffic knowledge injection, achieves an accuracy of 32.7\%, slightly surpassing that of random guessing (25\% accuracy).
Upon subjecting VIOLET to a single round of knowledge injection using the methodology proposed in \papername, we observe a significant improvement in accuracy, reaching 39.2\%. This improvement corresponds to a 6.5 percentage point increase in absolute value or a 19.88\% enhancement relative to the baseline performance.

Our analysis of the results reveals that the majority of these improvements are specifically attributed to the resolution of \textit{basic} questions within the \tqa dataset. 
These questions pertain to fundamental scene features, counting, and similar aspects. 
As elaborated in Section \ref{sec:methodology}, the generated captions produced by \papername primarily address this particular type of information.

In our future endeavors, we intend to broaden the scope of scene knowledge that can be captured and integrated within our framework. This expansion represents a critical stride towards further enhancing our approach and enabling advanced forms of reasoning, such as counterfactual inference, event forecasting, and more.

\subsection{Ethical Considerations}
Our research commenced with the utilization of publicly accessible video data, which was gathered through a combination of crowd contributions and rigorous neutralization processes. Despite our best efforts, it is essential to recognize that these data sets might still exhibit certain biases inherent to human perception and interpretation.
In order to extract meaningful information from the video data, we employed SOTA automatic annotation models. However, it is crucial to note that these models were primarily developed and optimized for traffic situations commonly observed in Western countries. As a consequence, the cultural context embedded within these models may introduce an additional layer of bias to the final results.
One specific manifestation of such bias could be the incomplete detection of all vehicles present in the video footage. Since the models were primarily trained and fine-tuned using data from Western countries, they might not perform as effectively in scenarios that deviate from the Western cultural context. Therefore, it is important to interpret the outputs of our models with caution, considering the potential limitations arising from cultural biases.

\subsection{Limitations of \papername}
The solution proposed in our study focuses specifically on addressing narrow traffic situations, as these scenarios present distinct challenges and considerations. 
By narrowing our scope to this domain, we were able to develop a solution that is tailored to the unique characteristics and requirements of such scenarios.

From a linguistic perspective, the generated sentences produced by our system exhibit a basic and rigid structure. While this simplicity may suffice for the immediate purpose of conveying information in the context of traffic situations, it is important to recognize that language generation from structured data is a well-researched area with extensive literature. This body of research offers various techniques and approaches that can be incorporated into future iterations of our work to enhance the complexity and flexibility of the generated sentences.  
Specifically, the introduction of large-scale language models, such as ChatGPT, which possess significant capabilities in generating high-quality text. Leveraging these state-of-the-art models opens up new possibilities for improving the quality and richness of the generated captions.

The quality of annotations can also limit the generalization of our work. 
As discussed in Section \ref{sec:methodology}, our method does not heavily rely on perfect quality for automatic annotations. 
While this flexibility allows for practical applications and reduces the dependency on precise annotations, it is crucial to recognize the potential challenges posed by adversarially noisy annotations. 
These noisy annotations, which may contain misleading or incorrect information, have the potential to negatively impact the performance of video-language models on the given task. 
Consequently, it is important for future research to delve deeper into the analysis and understanding of the effects of annotation quality on the performance and robustness of such models. 
By conducting thorough investigations into this aspect, we can gain valuable insights and inform the development of more resilient and accurate video-language models.

\section{CONCLUSIONS}
\label{sec:conclusion}
We propose \papername, a weak supervision method to inject traffic-domain knowledge into video-language models.
Our solution starts with raw traffic videos and uses a combination of automatic annotation tools and template-based language generation to create automatic captions for the video. 
We show that by fine-tuning a representative video-LM, VIOLET~\cite{fu2021violet}, on the combination of videos and generated captions, the model's performance improves in traffic-related reasoning tasks.
Our results on \tqa~\cite{trafficqa} benchmark show that the VIOLET's accuracy increases up to 6.5 absolute values in accuracy.

\bibliographystyle{plain}
\bibliography{refs}





\end{document}